\documentclass[runningheads]{llncs}
\usepackage{graphicx}
\usepackage{booktabs}
\usepackage{cite}
\usepackage{amsmath}
\usepackage[misc]{ifsym}
\makeatletter
\def\thanks#1{\protected@xdef\@thanks{\@thanks
		\protect\footnotetext{#1}}}
\makeatother

\begin{document}
\title{A Transformer-Based Adaptive Semantic Aggregation Method for UAV Visual Geo-Localization
\thanks{This work was supported in part by the National Natural Science Foundation of China (NSFC) under Grant No.62171295, and in part by the Liaoning Provincial Natural Science Foundation of China under Grant No.2021-MS-266, and in part by the Applied Basic Research Project of Liaoning Province under Grant 2023JH2/101300204, and in part by the Shenyang Science and Technology Innovation Program for Young and Middle-aged Scientists under Grant No.RC210427, and in part by the High Level Talent Research Start-up Fund of Shenyang Aerospace University under Grant No.23YB03.}
}
\titlerunning{Adaptive Semantic Aggregation Method for UAV Visual Geo-Localization}

\author{Shishen Li \and Cuiwei Liu \textsuperscript{\Letter} \and  Huaijun Qiu  \and Zhaokui Li }

\authorrunning{S. Li et al.}

\institute{School of Computer Science, Shenyang Aerospace University, Shenyang, China
	\email{liucuiwei@sau.edu.cn}}

\maketitle      
\begin{abstract}
This paper addresses the task of Unmanned Aerial Vehicles (UAV) visual geo-localization, which aims to match images of the same geographic target taken by different platforms, i.e., UAVs and satellites.
In general, the key to achieving accurate UAV-satellite image matching lies in extracting visual features that are robust against viewpoint changes, scale variations, and rotations.
Current works have shown that part matching is crucial for UAV visual geo-localization since part-level representations can capture image details and help to understand the semantic information of scenes.
However, the importance of preserving semantic characteristics in part-level representations is not well discussed.
%In this paper, we introduce a transformer-based adaptive semantic aggregation method that regards parts as the most representative semantics in an image and learns correlations of image patches to different parts in terms of the transformer's feature map.
In this paper, we introduce a transformer-based adaptive semantic aggregation method that regards parts as the most representative semantics in an image.
Correlations of image patches to different parts are learned in terms of the transformer's feature map.
Then our method decomposes part-level features into an adaptive sum of all patch features.
By doing this, the learned parts are encouraged to focus on patches with typical semantics.
Extensive experiments on the University-1652 dataset have shown the superiority of our method over the current works.

\end{abstract}
\section{Introduction}

Unmanned Aerial Vehicle (UAV) visual geo-localization refers to cross-view image retrieval between UAV-view images and geo-tagged satellite-view images.
Recently, this technology has been applied in many fields, such as precision agriculture~\cite{chivasa2020uav}, rescue system~\cite{rizk2021toward}, and environmental monitoring~\cite{ecke2022uav}.
Using a UAV-view image as the query, the retrieval system searches for the most relevant satellite-view candidate to determine the geographic location of the UAV-view target.
On the other hand, if a satellite-view image is used as the query, the corresponding UAV-view images can be retrieved from the UAV flight records, enabling UAV navigation.
Compared to traditional geo-localization methods that rely on GPS or radar, UAV visual geo-localization does not require the UAV to receive external radio information or emit detection signals, making it possible to achieve UAV positioning and navigation in radio silence.

The key to UAV visual geo-localization is to extract discriminative features.
%Considering different flight directions of drones, images taken in the same location may exhibit significant differences in content information, resulting in large intra-class distances between drone images and satellite images.
Specifically, satellites acquire fixed-scale images from a vertical view, while UAVs take images from various distances and orientations at an oblique view, resulting in large visual and scale variations in UAV-satellite image pairs.
%Moreover, images taken at different locations may have similar architectural styles as well as environmental appearances, resulting in small inter-class distances.
Moreover, images of different locations share some local patterns, such as vegetation and similarly styled buildings, which also pose challenges to cross-view image retrieval.
These issues make hand-crafted descriptors (e.g., SIFT~\cite{chiu2013fast}) perform poorly.

\begin{figure}[t]
	\centering
	\includegraphics[width=1\textwidth]{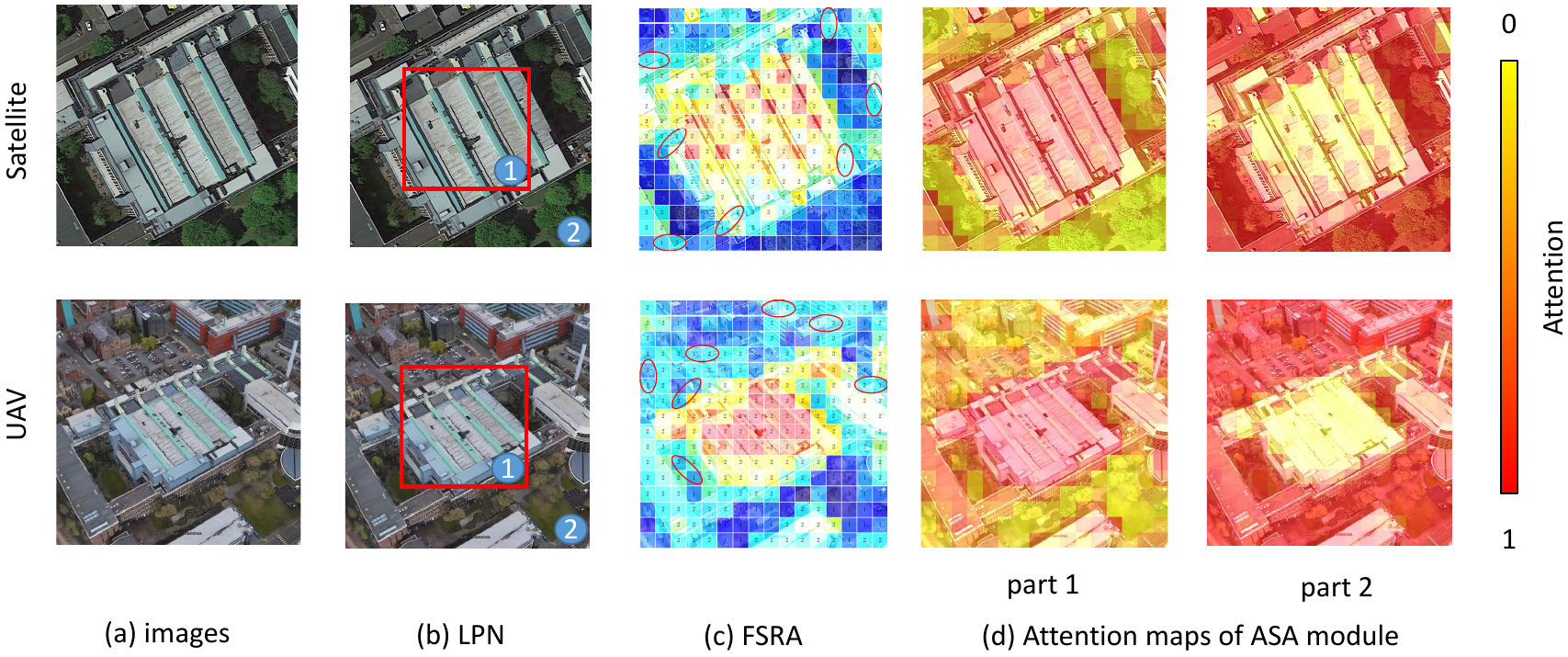}
	\vskip -5pt
	\caption{An UAV-satellite image pair is shown in column(a). The square-ring partition strategy of LPN~\cite{wang2021each} is depicted in column(b). Column(c) illustrates the heat map and the part partition of image patches generated by FSRA~\cite{dai2021transformer}. Red ellipses mark patches that are similar in features but divided into different parts. Attention maps corresponding to two parts produced by the proposed ASA module are given in column(d).}
	\label{Fig:partitions}
	\vskip -15pt
\end{figure}

Early UAV visual geo-localization methods employ two-branch CNN models~\cite{zheng2020university,ding2020practical} to achieve cross-view matching between UAV-view and satellite-view images.
Benefiting from the availability of multiple UAV-view images of the same location, the models are optimized in a location classification framework to learn view-invariant yet location-dependent features.
By doing this, two-branch CNN models are desired to generate similar features for satellite-view and UAV-view images taken at a new location that is unseen during training.
%With the development of deep learning, early works \cite{zheng2020university,ding2020practical} attempted to utilize deep neural networks to extract global features for cross-view image retrieval, achieving certain progress.
However, the learned features focus on the entire image, while neglecting fine-grained details that are crucial for distinguishing images of different locations.
Upon two-branch CNN models, Local Pattern Network (LPN)~\cite{wang2021each} divides the feature map into several regions with square-rings to extract part-level representations as shown in Fig.~\ref{Fig:partitions} (b).
Then part matching is performed to roughly align the geographic targets as well as their surroundings.
%LPN \cite{wang2021each} divides the feature maps into several different parts according to the distance to the image center and extracts multiple part-level features instead of global features, which can be considered as a spatial-based method.
Another iconic work is a two-branch transformer-based model called FSRA (Feature Segmentation and Region Alignment)~\cite{dai2021transformer}, which clusters image patches into semantic parts according to the heat distribution of feature maps as shown in Fig.~\ref{Fig:partitions} (c).
Each part is desired to indicate certain semantics such as target or surroundings.
Compared to LPN that adopts fixed-scale spatial partition, FSRA is more flexible in extracting parts and thus more robust against image shift and scale variations.
However, FSRA employs a hard partition strategy where an image patch belongs to only one part and the part-level representations are calculated as mean of image patches.
We argue that there are two issues pertaining to such strategy.
First, image patches similar in features may be divided into different parts as shown in Fig.~\ref{Fig:partitions} (c).
Secondly, such strategy cannot extract the most representative semantics since it neglects the associations between image patches and parts.

%Although current works~\cite{wang2021each,dai2021transformer} have demonstrated the effectiveness of part matching, they both employ hard partition strategies where an image patch (or a pixel) belongs to only one part and the part-level representations are calculated as mean of image patches (or pixels).

To cope with these limitations, this paper proposes an Adaptive Semantic Aggregation (ASA) module.
Unlike the hard partition strategy utilized in FSRA~\cite{dai2021transformer}, the ASA module employs a soft partition strategy that considers correlations between parts and all image patches to generate global-aware part-level representations.
Specifically, each part is regarded as one semantic and has a distribution of attention over all image patches as shown in Fig.~\ref{Fig:partitions} (d).
First, the most representative patch is selected as anchor of a part.
Attentions of image patches are allocated by calculating similarities between patches and the anchor.
A high attention expresses strong correlation of an image patch to the semantic part, while a low attention indicates weak correlation.
%Then representation of a part is calculated as the weighted summation of all patch features.
Then all patch-level features are adaptively aggregated into global-aware part-level representations according to the learned attentions.
Finally, the ASA module is integrated into the two-branch transformer-based framework~\cite{dai2021transformer} and explicitly enables the learned parts focus on distinctive patches, such as gray roof and circular road.

The remainder of this paper is organized as follows. 
Section 2 briefly describes the current work related to cross-view geo-localization. 
Section 3 introduces the overall framework of our method and describes the proposed ASA module in detail. 
Section 4 presents and analyzes the experimental results on the University-1652 dataset.
Section 5 summarizes this paper.

\section{Related Work}
%Cross-view geo-localization is first raised to deal with the ground-to-aerial geo-localization task, which retrieves geographic information of a ground-view query image by comparing it against geo-tagged aerial-view gallery images.
In 2020, Zheng et al.~\cite{zheng2020university} formulated the UAV visual geo-localization problem as bidirectional cross-view image retrieval between UAV-view and satellite-view images.
%They explained that UAV-to-satellite image retrieval can locate a UAV-view target while satellite-to-UAV image retrieval is able to achieve UAV navigation.
They employed a two-branch CNN model to extract global features from different domains and released the University-1652 dataset for model evaluation.
Upon this work, Ding et al.~\cite{ding2020practical} simplified the cross-view image retrieval task as a location classification problem during training, aiming to learn a common location-dependent feature space that is well scalable for unseen images.
They noticed the imbalance of UAV-view and satellite-view images, and performed data augmentation to expand training satellite-view images.
Based on the above location classification framework, recent works~\cite{wang2021each,tian2021uav,zhuang2021faster,lin2022joint,dai2021transformer,zhuang2022semantic} have made various attempts to improve the discriminative power of the learned feature space.

One typical solution is LPN~\cite{wang2021each} which achieves fine-grained part matching between UAV-view and satellite-view images. 
LPN applies a square-ring partition strategy to separate global feature maps into multiple parts according to their spatial position.
Then rotation-invariant part-level features are obtained by performing average polling over points within each part. 
Tian et al.~\cite{tian2021uav} produces synthesized UAV-view images by a generative model to reduce the gap between two views.
Then they employed LPN~\cite{wang2021each} to achieve cross-view image retrieval.
Zhuang et al.~\cite{zhuang2021faster} improved LPN~\cite{wang2021each} by incorporating global features and adding KL loss to further close the distance between paired UAV-view and satellite-view images.
Lin et al.~\cite{lin2022joint} introduced a Unit Subtraction Attention Module (USAM) that forces the geo-localization model (e.g., LPN~\cite{wang2021each}) to focus on salient regions by detecting representative key points.

Considering that the square-ring partition strategy is not robust to scale variations, Dai et al.~\cite{dai2021transformer} aimed to extract semantic parts composed of patches scattered throughout the image.
Small patches are ranked according to the feature heat map and uniformly divided into multiple parts, regardless of their spatial position.
Then a part is represented by the average feature of patches within it.
Zhuang et al.~\cite{zhuang2022semantic} improved this part partition strategy by searching for the optimal split based on the gradient between adjacent positions in the ranking results.
However, the above methods~\cite{dai2021transformer,zhuang2022semantic} aggregate image patches equally, thus weakening the semantic characteristics of the learned parts.
In this paper, we propose an Adaptive Semantic Aggregation module that regards parts as the most representative semantics in images and obtains part-level features by aggregating all patch features based on their correlations to parts.

\begin{figure}[t]
	\includegraphics[width=\textwidth]{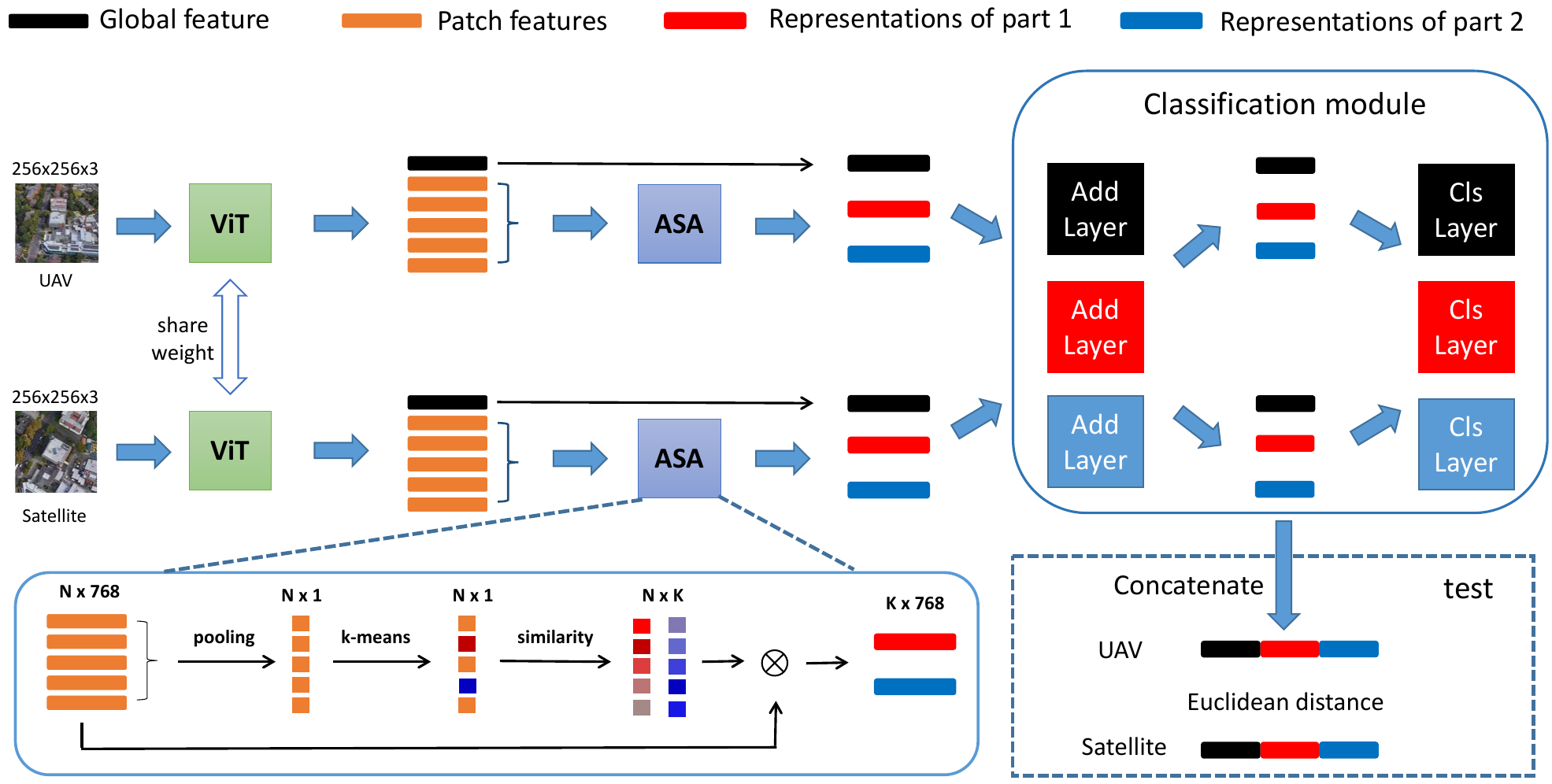}
	\vskip -5pt
	\caption{Overall framework of our method.}
	\vskip -15pt
	\label{Fig:framework}
\end{figure}

\section{Method}
Fig.~\ref{Fig:framework} depicts the overall framework of our method.
Inspired by siamese networks\cite{bromley1993signature,koch2015siamese}, two branches are designed to handle images captured by UAVs and satellites respectively.
Following FSRA~\cite{dai2021transformer}, we adopt ViT (Vision Transformer)~\cite{dosovitskiy2020image} as the backbone to extract features from both UAV-view images and satellite-view images.
Backbones of the UAV branch and the satellite branch share weights to learn a mapping function that is able to project images from both views to one common embedding space.
ViT produces global features of the entire image as well as local features of image patches that are fed into the proposed ASA module to generate multiple part-level representations. 
Finally, global features and part-level representations are sent to the classification module, which regards geographic locations of training images as semantic categories.
The classification module contains additive layers for representation transformation and classification layers for geo-location prediction.

In the test stage, our goal is to achieve cross-view retrieval between UAV-view and satellite-view images captured at new geographic locations.
That is to say, the classifier cannot infer locations of query or gallery images during test, since the test data have their own location label space disjoint with the training data.
Therefore, we concatenate the transformed representations before classification layers in the classifier module as the final descriptor of a test image.
In order to measure the correlation between a query image from one view and gallery images from another view, we calculate the euclidean distance between them.
Finally, we sort gallery images in terms of their distance to the query image and return the most similar one to achieve cross-view image retrieval.

\begin{figure}[t]
	\centering
	\includegraphics[width=0.75\textwidth,height=0.36\textwidth]{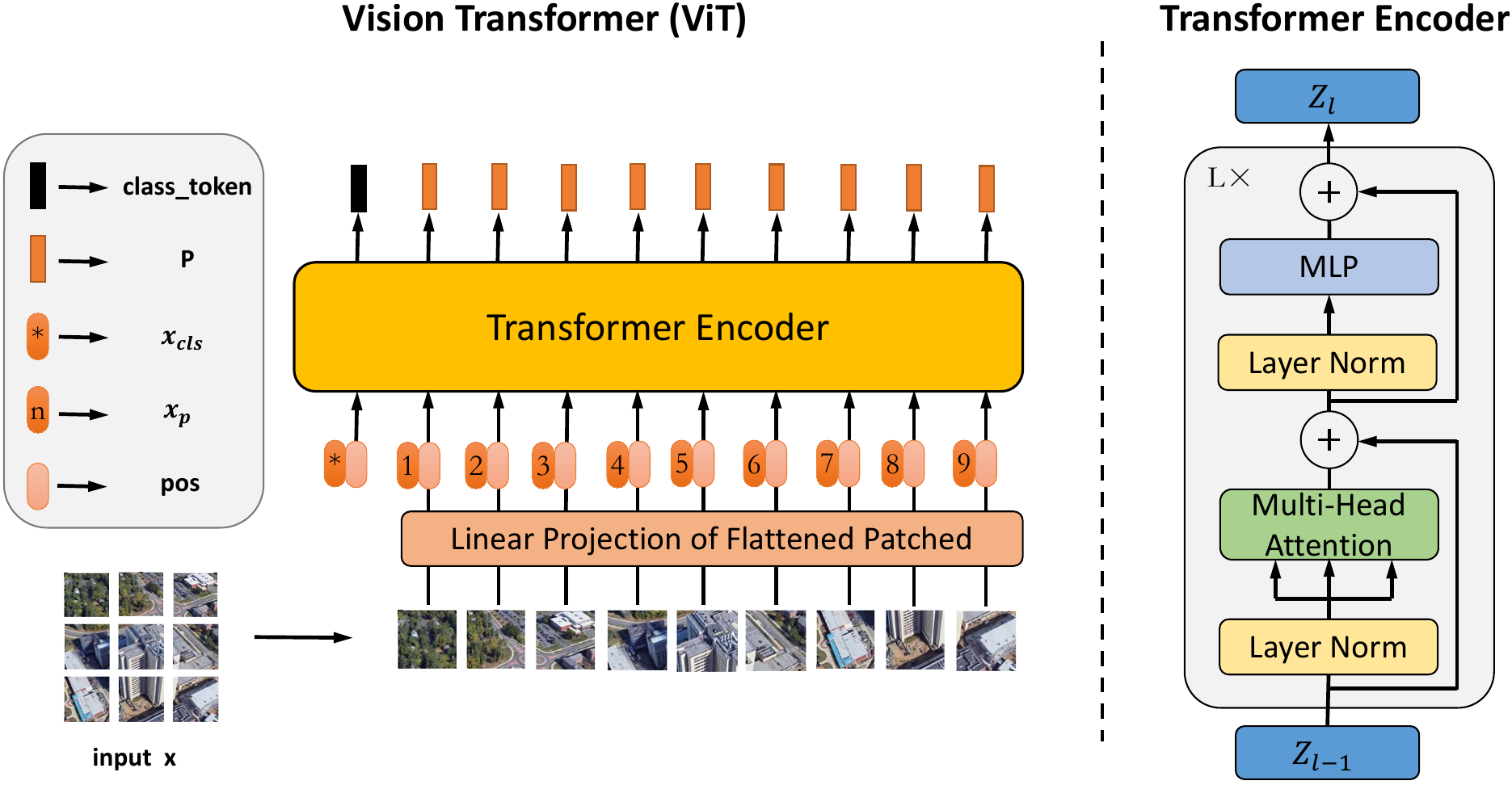}
	\vskip -5pt
	\caption{Architecture of Vision Transformer (ViT).}
	\vskip -10pt
	\label{Fig:Vit}
\end{figure}

\subsection{Transformer-based backbone}
Architecture of our backbone is illustrated in Fig.~\ref{Fig:Vit}.
Given an input image $x \in R^{H \times W \times C}$, it is first divided into fixed-size patches using pre-defined parameters.
Then patches are linearly transformed to obtain embedding vectors $x_p \in R^{N \times D}$, where $N$ and $D$ refer to the number of patches and the embedding dimension, respectively.
Additionally, a learnable vector $x_{cls} \in R^{1 \times D}$ forwards along with $x_p$.
All these embedding vectors are integrated with position embeddings $pos \in R^{(N+1)\times D}$ to obtain input vectors $Z_0$ of the Transformer Encoder.
This procedure is formulated by
\begin{equation}
	Z_0 = [x_{cls}; x_p] + pos.
\end{equation}
To accommodate the varying resolution of input images, we employ learnable position embeddings instead of utilizing parameters pre-trained on ImageNet~\cite{deng2009imagenet}.

As shown in Fig.~\ref{Fig:Vit}, the Transformer Encoder consists of alternating Multiple Head Self-Attention layers and Layer Norm (LN) operations, where residual connections are applied. 
Multi-Layer Perceptron (MLP) is a two-layer non-linearity block, each layer of which ends with a GELU activation function. 
The process of Transformer Encoder is formulated by
\begin{equation}
	Z^{'}_{l} = MHSA(LN(Z_{l-1})) + Z_{l-1},
\end{equation}
\begin{equation}
	Z_l = MLP(LN(Z^{'}_{l})) + Z^{'}_{l},
\end{equation}
where $Z^{'}_{l}$ and $Z_l$ denote output vectors of the $l^{th}$ attention layer and the $l^{th}$ MLP layer, respectively.
Finally, the output vector $Z_L$ consists of a global feature $class\_token\in R^{1 \times D}$ generated upon $x_{cls}$ and patch-level features $\{P_i \in R^{1 \times D}\}_{i=1:N}$ generated from $x_p$.

\subsection{Adaptive semantic aggregation module}

%Previous works have already shown the effectiveness of the Transformer in UAV-view geo-localization task. However, challenges such as position shift and scale adaption still need to be addressed\cite{dai2021transformer}.
Previous works have shown the effectiveness of part matching between UAV-view and satellite-view images in the UAV-view geo-localization task.
In this paper, we propose a soft partition strategy to adaptively aggregate image patches into part-level representations.
Different from hard partition strategies~\cite{wang2021each,dai2021transformer}, the proposed soft partition strategy regards a part as semantic aggregation of image patches according to correlations between them.
Details of the ASA module are illustrated in Fig.~\ref{Fig:framework}.

An input image $x$ is first fed into ViT to extract global features as well as patch-level features $\{P_i \in R^{1 \times D}\}_{i=1:N}$.
Assuming equal importance of feature dimensions, we apply average pooling on features of each patch to obtain 1-dimensional representations, denoted as $\{Q_i \in R^{1 \times 1}\}_{i=1:N}$.
\begin{equation}
    Q_i = \frac{1}{D} \sum_{d=1}^D P_i^d,
    \label{Eq:semantic}
\end{equation}
where i and d represent the indices of patch and feature dimension, respectively. 
Patches with similar semantic information exhibit similar representations in $\{Q_i\}_{i=1:N}$, as demonstrated in FSRA~\cite{dai2021transformer} (see Fig.~\ref{Fig:partitions} (c)).
%However, Fig.~\ref{Fig:partitions} (c) also shows that some patches with similar representations may be allocated to different parts by the hard partition strategy adopted in FSRA.
Accordingly, we perform k-means algorithm upon $\{Q_i\}_{i=1:N}$ to extract representative semantics.

Specifically, the 1-dimensional representations are first sorted in descending order to obtain a sequence $S$, each item in which indicates a patch index.
We introduce a hyper-parameter $K$, which denotes the number of semantic categories in the image. 
Center of the $k^{th}$ category is initialized as 1-dimensional representations of a patch $S_{IC_k}$, and the index $IC_k$ is denoted by
\begin{equation}
	IC_k = \frac{(2k-1)N}{2K}.   \qquad k = {1,2,...K}
\end{equation}
The k-means algorithm updates semantic categories iteratively and outputs the final category centers.
Center of the $k^{th}$ category corresponds to a patch $P_{C_k}$, which is regarded as the anchor of a part.
Then we utilize the original patch features produced by ViT to calculate Euclidean distance $dis_k^i$ between anchor of a part and all image patches.
This procedure is formulated by
\begin{equation}
    dis_k^i = ||P_i-P_{C_k}||_2.
\end{equation}
Given the negative correlation between distance and similarity, we employ the cosine function to obtain the attention matrix by
\begin{equation}
    A_k^i = \alpha \cdot cos(\frac{dis_k^i - dis_k^{min}}{dis_k^{max} - dis_k^{min}} \cdot \frac{\pi}{2}) + \beta,
\end{equation}
where $\alpha$ and $\beta$ are factors for enhancing the robustness of the weight matrix and are respectively set to 1 and 0 in our specific experiments.
Finally, we aggregate patch features $\{P_i\}_{i=1:N}$ into global-aware part-level features $\{\rho_k\}_{k=1:K}$ according to the customized attention matrix $A$.
\begin{equation}
    \rho_k = \frac{\sum_{i=1}^{N}P_i \cdot A_k^i}{\sum_{i=1}^{N}A_k^i}.
\end{equation}

%It should be noted that during the feature aggregation process, we use the original D-dimensional features $\{P_i\}_{i=1:N}$ rather than the 1-dimensional semantic representations $\{Q_i\}_{i=1:N}$.
It should be noted that 1-dimensional semantic representations $\{Q_i\}_{i=1:N}$ are utilized to find anchors of parts at low clustering cost, while original D-dimensional features $\{P_i\}_{i=1:N}$ are employed for feature aggregation.
In fact, clustering results of the k-means algorithm can also be used to achieve a hard partition strategy, which is compared to the proposed soft partition strategy on the University-1652 dataset in Section 4.

\subsection{Classification module}
\begin{figure}[t]
	\centering
	\includegraphics[width=1\textwidth]{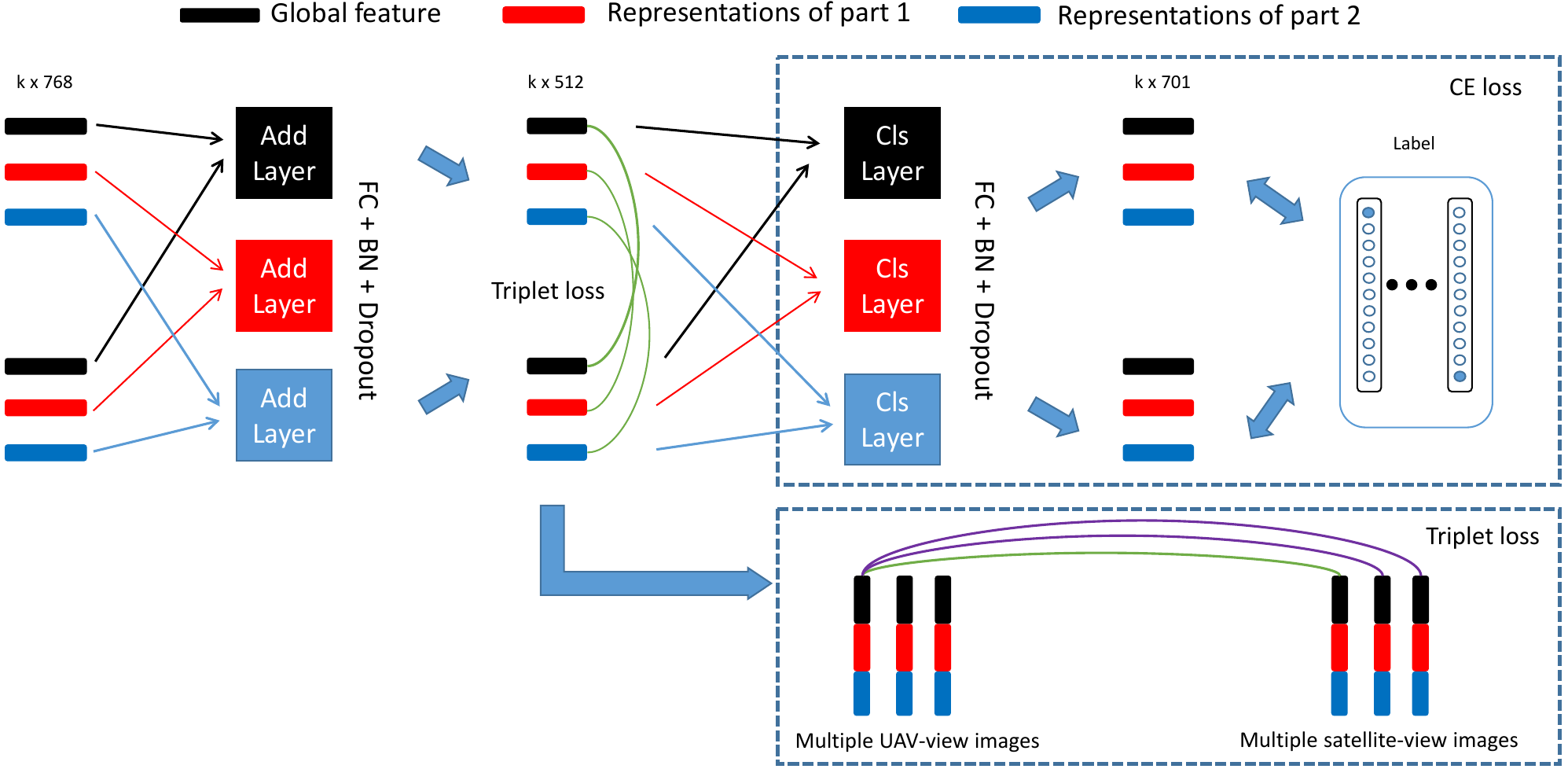}
	\vskip -5pt
	\caption{Architecture of the classification module. In training, global and part-level features are fed into additive layers followed by classification layers. Suppose that the training data come from 701 locations, so a classification layer predicts a 701-dimensional vector. The model is optimized by CE loss and Triplet loss. Green and purple lines point at positive and negative samples for calculating the Triplet loss, respectively.}
	\vskip -10pt
	\label{Fig:cllassifier}
\end{figure}

The classification module takes global features as well as part-level representations as input, aiming at classifying them into different geographic locations.
As shown in Fig.~\ref{Fig:cllassifier}, the classification module consists of additive layers and classification layers.
The former achieve transformation of the input while the latter produce prediction vectors of geographic locations.
Separate layers are constructed for global features and representations of each part, considering that they indicate different semantic characteristics.
Take representations of the $k^{th}$ part as an example, operations of the classification module can be formulated as
\begin{equation}
	f_k = F_{add}^k(\rho_k),
\end{equation}
\begin{equation}
	z_k = F_{cls}^k(f_k),
\end{equation}
where $F_{add}^k$ and $F_{cls}^k$ represents the additive layer and the classification layer for the $k^{th}$ part, respectively.
$f_k$ denotes output features of the additive layer and $z_k$ indicates predicted logits for all geographic locations.

Following~\cite{dai2021transformer}, the model is optimized by both CE loss $L_{CE}$ and Triplet Loss $L_{Triplet}$.
The total objective is defined as
\begin{equation}
	L_{total} = L_{CE} + L_{Triplet},
\end{equation}
where $L_{CE}$ is the cross entropy between the predicted logits and the ground-truth geo-tags, formulated by
\begin{equation}
	L_{CE} = -\frac{1}{K}\sum_{i=1}^{K}\log\frac{exp(z_k(y))}{\sum_{c=1}^{C}exp(z_k(c))},
\end{equation}
where $C$ indicates the number of geographic locations in the training data and $z_k(y)$ is the logit score of the ground-truth geo-tag $y$. 
The Triplet Loss is performed on the output features of additive layers to pull paired UAV-view and satellite-view images together, while pushing away mismatched image pairs.
The Triplet Loss is formulated by
\begin{equation}
	L_{TripletLoss} = \frac{1}{K}\sum_{i=1}^{K}max(d(f_k,p)-d(f_k,n)+M;0),
\end{equation}
Where $M$ is a hyper-parameter and empirically set to 0.3.
$d(.)$ represents the distance function.
$p$ and $n$ indicate positive sample and negative samples, respectively.
As depicted in Fig.~\ref{Fig:cllassifier}, both the positive (green line) and negative samples (purple lines) come from another view.
For example, if we take part-level feature $f_k$ of a UAV-view image as the anchor, then $f_k$ is compared to the corresponding part features of the true-matched and mismatched satellite-view images.

\section{Experiments}
\subsection{Dataset}
%University-1652 dataset~\cite{zheng2020university} is a widely used multi-view multi-source benchmark that contains images of 1652 buildings from 72 universities in three different camera views, namely UAV view, satellite view, and ground view.
University-1652 dataset~\cite{zheng2020university} contains images of 1652 buildings from 72 universities in three different camera views, i.e. UAV view, satellite view, and ground view.
This dataset is employed to evaluate the proposed method on the UAV visual geo-localization task, so only UAV-view and satellite-view images are utilized to achieve bidirectional cross-view image retrieval.
For each building, one image was obtained by satellites and 54 images were captured by UAVs at different heights and perspectives.
%For each building, one satellite-view image as well as 54 UAV-view images taken at different heights and perspectives. 
Thus, there exist large variations in image scale and viewpoints, which poses great challenges to the UAV visual geo-localization task.

The dataset is split into a training set including 38,555 images of 701 buildings from 33 universities and a test set containing 52,306 images of 951 buildings from the reminder 39 universities.
%Obviously, the test images for cross-view image retrieval have their own location label space independent to the training images.
During training, we number 701 buildings into 701 different categories of the classification module and learn an embedding space for UAV-view and satellite-view images.
In the UAV-to-satellite image retrieval task, the query set consists of 37,855 UAV-view images of 701 buildings and the gallery includes 701 true-matched satellite-view images and 250 distractors.
In the satellite-to-UAV image retrieval task, 701 satellite-view images constitute the query set and there are 51,355 gallery images, including 37,855 true-matched UAV-view images and 13,500 distractors from the rest 250 buildings.

\subsection{Evaluation protocol}
%We apply Recall@K (R@K) and average precision (AP) as evaluation metrics to achieve fair comparison with the previous works. 
%The value of Recall@K is set to 1 when there exists at least one true-matched gallery image in the first K retrieved images.
%Recall@K is calculated by
%\begin{equation}
%	Recall@K =
%	\begin {cases} 1, &r_{true} < K+1 \\
%	0, &otherwise\\
%	\end {cases}
%\end{equation}
%where $r_{true}$ is order of the first true-matched gallery image in the ranking list.
%It is clear that Recall@K only cares about whether the system can return a true-match gallery image in the first K retrieved results.
%Therefore, this metric is suitable for settings where there is only one true match image.

Following the previous works~\cite{zheng2020university,ding2020practical,wang2021each,zhuang2021faster,tian2021uav}, we employ two types of evaluation metrics, namely Recall@K and average precision (AP).
Recall@K and AP are widely applied in image retrieval tasks.
Recall@K refers to the ratio of query images with at least one true-matched image appearing in the top-K ranking list.
AP computes the area under the Precision-Recall curve and considers all true-matched images in the gallery.
Recall@K focuses on the position of the first true-matched image in the matching results and thus is suitable for evaluation on the UAV-to-satellite image retrieval task where the gallery only contains one true-matched satellite-view image for each UAV-view query.
In the satellite-to-UAV image retrieval task, there are multiple true-matched UAV-view images for one satellite-view query and AP is able to comprehensively measure the matching results.
In this paper, we report the mean Recall@K and mean AP of all queries.

\subsection{Implementation details}
%Our method is produced in the Pytorch~\cite{paszke2019pytorch} framework and all the experiments are executed on one Nvidia GTX 3070 GPU.
All the input images are resized to $ 256 \times 256 $ and the number of parts ($K$) is set to 2. 
Due to imbalance in the number of UAV-view and satellite-view images, we perform image augmentation on satellite-view images during training.
A small Vision Transformer (ViT-S) pre-trained on the ImageNet~\cite{deng2009imagenet} is employed as the backbone for feature extraction.
The ASA module and the classification module are trained from scratch.
Our model is learned by an SGD optimizer with a momentum of 0.9 and weight decay of 0.0005.
The mini-batch size is set to 8.
The learning rate is initialized to 0.003 and 0.01 for ViT-S and the rest layers, respectively.
The model is trained for a total of 120 epochs and the learning rate is decayed by 0.1 after executing 70 epochs and 110 epochs. 

\begin{table} [t]
	\centering
	\caption{Cross-view image retrieval performance (\%) of different methods.}
		\label{Tab:comparison}
		\vskip -5pt
	\begin{tabular}{c|c|c|cc|cc}
		\toprule
		Method & Year & Backbone & \multicolumn{2}{|c|}{UAV-to-Satellite}  & \multicolumn{2}{|c}{Satellite-to-UAV} \\
		&	&  & Recall@1   &AP & Recall@1   &AP \\
		\hline
%		Soft margin triplet loss~\cite{hu2018cvm}  &2020 & VGG16& 53.21 &58.03& 65.62 &54.47	\\
		Zheng et al.~\cite{zheng2020university} &2020 & ResNet-50	&58.49 &63.31 &71.18 &58.74	\\
		LCM~\cite{ding2020practical}  &2020	& ResNet-50	&  66.65 & 70.82& 79.89 &65.38			\\
		LPN~\cite{wang2021each} &2021	& ResNet-50	& 75.93 & 79.14&  86.45 &74.79			\\
		PCL~\cite{tian2021uav}  &2021  & ResNet-50	& 79.47 & 83.63&  87.69 &78.51			\\
		%MSBA\cite{zhuang2021faster} &2021 & ResNet-50	& 82.33 & 84.78&  90.58 &81.61			\\
		RK-Net (USAM)~\cite{lin2022joint} &2022 & ResNet-50 &77.60	&80.55	&86.59	&75.96      \\
		SGM~\cite{zhuang2022semantic}  & 2022 & Swin-T    & 82.14 & 84.72&  88.16 &81.81          \\
		FSRA~\cite{dai2021transformer} &2022 & Vit-S		& 84.51 & 86.71&  88.45 &83.37			\\
		\hline
		ASA (Ours) &--  & Vit-S 	& 85.12 &87.21& 89.30 &84.17			\\
		\bottomrule
	\end{tabular}
\vskip -5pt
\end{table}

\begin{table} [t]
	\centering
	\caption{Cross-view image retrieval performance (\%) with different strategies.}
	\label{Tab:partition}
	\vskip -5pt
	\begin{tabular}{c|cc|cc}
		\toprule
		Strategy & \multicolumn{2}{|c|}{UAV-to-Satellite}  & \multicolumn{2}{|c}{Satellite-to-UAV} \\
		&Recall@1   &AP & Recall@1   &AP \\
		%		\hline
		%		ViT-S(backbone) & 71.04 & 74.62 & 83.31 & 72.08		\\
		\hline
		Uniform hard partition strategy &83.98  &86.27 & 88.59  &	83.91\\
		k-means	hard partition strategy & 84.97 & 87.12 & 88.16 & 83.85			\\
		k-means	soft partition strategy & 85.12 & 87.21 & 89.30 & 84.17			\\
		\bottomrule
	\end{tabular}
	\vskip -10pt
\end{table}

\subsection{Experimental results}
\subsubsection{Comparison with existing methods.}
Table~\ref{Tab:comparison} presents our model's performance for bidirectional cross-view retrieval between UAV-view and satellite-view images, compared to the previous methods~\cite{zheng2020university,ding2020practical,wang2021each,tian2021uav,lin2022joint,zhuang2022semantic,dai2021transformer}.
LCM~\cite{ding2020practical} resizes input images into $384 \times 384$, and the image scale utilized in the other methods is $256 \times 256$.
Methods in~\cite{zheng2020university,ding2020practical,wang2021each,tian2021uav,lin2022joint} all use ResNet-50~\cite{he2016deep} as backbone to extract CNN features.
Among them, methods in~\cite{zheng2020university,ding2020practical} learn global features, while LPN~\cite{wang2021each}, PCL~\cite{tian2021uav}, and RK-Net~\cite{lin2022joint} achieve part matching by using the square-ring partition strategy, which is not robust to scale variations.
The experimental results demonstrate the inferior performance of these methods~\cite{zheng2020university,ding2020practical,wang2021each,tian2021uav,lin2022joint}.
FSRA~\cite{dai2021transformer} and SGM~\cite{zhuang2022semantic} adopt transformers as backbone and cluster patches into semantic parts with hard partition strategies.
Our method performs better than FSRA~\cite{dai2021transformer} and SGM~\cite{zhuang2022semantic} on both tasks, even though SGM~\cite{zhuang2022semantic} employs a stronger backbone.
In comparison to FSRA~\cite{dai2021transformer}, both methods extract features with ViT-S backbone and utilize $3 \times$ sampling strategy to expand satellite-view images during training.
The superior results achieved by our method demonstrate that the proposed ASA module reasonably aggregate patches into part-level representations, thus improving the performance of part matching.

\begin{table}[t]
	\centering
	\caption{Effect of the number of parts on cross-view image retrieval performance (\%).}
	\label{Tab:numofparts}
	\vskip -5pt
	\begin{tabular}{c|cc|cc}
		\toprule
		Number of parts ($K$) & \multicolumn{2}{|c|}{UAV-to-Satellite}  & \multicolumn{2}{|c}{Satellite-to-UAV} \\
		&Recall@1   &AP & Recall@1   &AP \\
		\hline
		1   & 72.11 & 75.59 & 79.46 & 71.83 		\\
		2	& 85.12 & 87.21 & 89.30 & 84.17	        \\
		3   & 84.73 & 86.88 & 88.45 & 83.55 			\\
		4	& 84.48 & 86.74 & 87.59 & 83.22			\\
		\bottomrule
	\end{tabular}
	\vskip -5pt
\end{table}

\begin{table}[t] 
	\centering
	\caption{Cross-view image retrieval performance (\%) with different image sizes.}
	\label{Tab:size}
	\vskip -5pt
	\begin{tabular}{c|cc|cc}
		\toprule
		Image Size & \multicolumn{2}{|c|}{UAV-to-Satellite}  & \multicolumn{2}{|c}{Satellite-to-UAV} \\
		&Recall@1   &AP & Recall@1   &AP \\
		\hline
		$224 \times 224$   &82.28  &84.83  &86.31  &81.93 		\\
		$256 \times 256$   & 85.12 & 87.21 & 89.30 & 84.17	\\
		$384 \times 384$   &86.88  &88.74  &89.44  &85.95 			\\
		$512 \times 512$   & 88.67 & 90.29 &89.44  &87.14 			\\
		\bottomrule
	\end{tabular}
	\vskip -10pt
\end{table}

\subsubsection{Ablation study.}
In this section, we investigate the effect of several key factors.
First, we explore the effectiveness of the proposed soft partition strategy by comparing it with two baselines in Table~\ref{Tab:partition}.
The baseline ``Uniform hard partition strategy" uniformly groups patches into $K$ parts according to the 1-dimensional semantic representation obtained by Eq.~\ref{Eq:semantic}.
The baseline ``k-means hard partition strategy" utilizes the k-means clustering results for part partition.
Both baselines take the average of patches within each part to generate part-level representations.
As shown in Table~\ref{Tab:partition}, the proposed ``k-means soft partition strategy" is more effective in aggregating part-level features and achieves optimal results for UAV visual geo-localization.

Next, we analyze the impact of the number of parts $K$ on the cross-view image retrieval performance.
As shown in Table~\ref{Tab:numofparts}, the performance is poor if K is set to 1, since the learned part cannot express representative semantics of the image.
Our model achieves the optimal generalization performance when K is set to 2.
We believe that the learned two parts indicate typical semantics of foreground and background.

Finally, we discuss the effect of different image sizes.
As shown in Table~\ref{Tab:size}, high-resolution images generally yield better results since they retain more fine-grained details at the cost of more memory resources and inference time.
The performance of our method does not degrade significantly when the image size is reduced from 512 to 256. 
Therefore, we recommend using the image size of 256 if hardware resources are limited.

\section{Conclusion}
%Part matching between UAV-view and satellite-view images has proved to be critical in UAV visual geo-localization task.
In this paper, we have presented an adaptive semantic aggregation method to learn global-aware part-level features that can express typical semantics of scenes.
Unlike the existing hard partition strategies, we have developed a soft partition strategy that searches for the most representative semantics as parts and evaluates the significance of patches to different parts.
Our method adaptively aggregates features of all patches into part-level representations.
Compared to current works, the proposed method has achieved superior performance on the University-1652 dataset.
Ablation studies also verified effectiveness of the proposed soft partition strategy and investigated several key factors of our methods.

% ---- Bibliography ----
%
% BibTeX users should specify bibliography style 'splncs04'.
% References will then be sorted and formatted in the correct style.
%
\bibliographystyle{splncs04_unsort}
\bibliography{samplepaper}

\begin{thebibliography}{10}
\providecommand{\url}[1]{\texttt{#1}}
\providecommand{\urlprefix}{URL }
\providecommand{\doi}[1]{https://doi.org/#1}

\bibitem{chivasa2020uav}
Chivasa, W., Mutanga, O., Biradar, C.: Uav-based multispectral phenotyping for
  disease resistance to accelerate crop improvement under changing climate
  conditions. Remote sensing  \textbf{12}(15), ~2445 (2020)

\bibitem{rizk2021toward}
Rizk, M., Slim, F., Charara, J.: Toward ai-assisted uav for human detection in
  search and rescue missions. In: DASA. pp. 781--786. IEEE (2021)

\bibitem{ecke2022uav}
Ecke, S., Dempewolf, J., Frey, J., Schwaller, A., Endres, E., Klemmt, H.J.,
  Tiede, D., Seifert, T.: Uav-based forest health monitoring: A systematic
  review. Remote Sensing  \textbf{14}(13),  3205--3249 (2022)

\bibitem{chiu2013fast}
Chiu, L.C., Chang, T.S., Chen, J.Y., Chang, N.Y.C.: Fast sift design for
  real-time visual feature extraction. TIP  \textbf{22}(8),  3158--3167 (2013)

\bibitem{wang2021each}
Wang, T., Zheng, Z., Yan, C., Zhang, J., Sun, Y., Zheng, B., Yang, Y.: Each
  part matters: Local patterns facilitate cross-view geo-localization. TCSVT
  \textbf{32}(2),  867--879 (2021)

\bibitem{dai2021transformer}
Dai, M., Hu, J., Zhuang, J., Zheng, E.: A transformer-based feature
  segmentation and region alignment method for uav-view geo-localization. TCSVT
   \textbf{32}(7),  4376--4389 (2022)

\bibitem{zheng2020university}
Zheng, Z., Wei, Y., Yang, Y.: University-1652: A multi-view multi-source
  benchmark for drone-based geo-localization. In: ACM MM. pp. 1395--1403 (2020)

\bibitem{ding2020practical}
Ding, L., Zhou, J., Meng, L., Long, Z.: A practical cross-view image matching
  method between uav and satellite for uav-based geo-localization. Remote
  Sensing  \textbf{13}(1), ~47 (2020)

\bibitem{tian2021uav}
Tian, X., Shao, J., Ouyang, D., Shen, H.T.: Uav-satellite view synthesis for
  cross-view geo-localization. TCSVT  \textbf{32}(7),  4804--4815 (2021)

\bibitem{zhuang2021faster}
Zhuang, J., Dai, M., Chen, X., Zheng, E.: A faster and more effective
  cross-view matching method of uav and satellite images for uav
  geolocalization. Remote Sensing  \textbf{13}(19), ~3979 (2021)

\bibitem{lin2022joint}
Lin, J., Zheng, Z., Zhong, Z., Luo, Z., Li, S., Yang, Y., Sebe, N.: Joint
  representation learning and keypoint detection for cross-view
  geo-localization. TIP  \textbf{31},  3780--3792 (2022)

\bibitem{zhuang2022semantic}
Zhuang, J., Chen, X., Dai, M., Lan, W., Cai, Y., Zheng, E.: A semantic guidance
  and transformer-based matching method for uavs and satellite images for uav
  geo-localization. IEEE Access  \textbf{10},  34277--34287 (2022)

\bibitem{bromley1993signature}
Bromley, J., Guyon, I., LeCun, Y., S{\"a}ckinger, E., Shah, R.: Signature
  verification using a" siamese" time delay neural network. Neurips  \textbf{6}
  (1993)

\bibitem{koch2015siamese}
Koch, G., Zemel, R., Salakhutdinov, R., et~al.: Siamese neural networks for
  one-shot image recognition. In: ICML deep learning workshop. vol.~2. Lille
  (2015)

\bibitem{dosovitskiy2020image}
Dosovitskiy, A., Beyer, L., Kolesnikov, A., Weissenborn, D., Zhai, X.,
  Unterthiner, T., Dehghani, M., Minderer, M., Heigold, G., Gelly, S., et~al.:
  An image is worth 16x16 words: Transformers for image recognition at scale.
  arXiv preprint arXiv:2010.11929  (2020)

\bibitem{deng2009imagenet}
Deng, J., Dong, W., Socher, R., Li, L.J., Li, K., Fei-Fei, L.: Imagenet: A
  large-scale hierarchical image database. In: CVPR. pp. 248--255. IEEE (2009)

\bibitem{he2016deep}
He, K., Zhang, X., Ren, S., Sun, J.: Deep residual learning for image
  recognition. In: CVPR. pp. 770--778 (2016)

\end{thebibliography}

\end{document}